\documentclass{article} 
\usepackage{iclr2025_conference,times}

\usepackage{amsmath,amsfonts,bm}

\def\eqref#1{equation~\ref{#1}}

\def\1{\bm{1}}

\DeclareMathAlphabet{\mathsfit}{\encodingdefault}{\sfdefault}{m}{sl}
\SetMathAlphabet{\mathsfit}{bold}{\encodingdefault}{\sfdefault}{bx}{n}

\usepackage{hyperref}
\usepackage{url}
\usepackage{enumitem}
\usepackage{todonotes}
\usepackage{wrapfig}
\usepackage{subcaption}

\title{Guided Sequence-Structure Generative \\Modeling for Iterative Antibody Optimization}

\author{Aniruddh Raghu\\
BigHat Biosciences\\
\texttt{araghu@bighatbio.com}
\And
Sebastian Ober\\
BigHat Biosciences\\ 
\And 
Maxwell Kazman\thanks{Work done while at BigHat Biosciences}\\
University of Washington\\
\And
Hunter Elliott\\ 
BigHat Biosciences\\
}

\iclrfinalcopy 
\begin{document}

\maketitle

\begin{abstract}
Therapeutic antibody candidates often require extensive engineering to improve key functional and developability properties before clinical development. 
This can be achieved through iterative design, where starting molecules are optimized over several rounds of \textit{in vitro} experiments.
While protein structure can provide a strong inductive bias, it is rarely used in iterative design due to the lack of structural data for continually evolving lead molecules over the course of optimization. 
In this work, we propose a strategy for iterative antibody optimization that leverages both sequence and structure as well as accumulating lab measurements of binding and developability.
Building on prior work, we first train a sequence-structure diffusion generative model that operates on antibody-antigen complexes. 
We then outline an approach to use this model, together with carefully \textit{predicted} antibody-antigen complexes, to optimize lead candidates throughout the iterative design process. 
Further, we describe a guided sampling approach that biases generation toward desirable properties by integrating models trained on experimental data from iterative design.
We evaluate our approach in multiple \textit{in silico} and \textit{in vitro} experiments, demonstrating that it produces high-affinity binders at multiple stages of an active antibody optimization campaign.
\end{abstract}
\section{Introduction}
Therapeutic antibodies are a flexible and rapidly-growing class of drugs that have already successfully been used to treat a wide range of diseases \citep{carter2018next}. However, developing these drugs is a challenging multi-objective protein engineering problem, since viable clinical candidates must not only bind their targets but also meet a range of developability criteria such improved stability and half-life \citep{jarasch2015developability}.

ML-based approaches can be used to engineer antibodies for desired properties of interest. In particular, active learning frameworks that iteratively refine candidates through repeated rounds of \textit{in vitro} validation have demonstrated value in prior work \citep{stanton2022accelerating,gruver2024protein,amin2024bayesian} (related work in Appendix \ref{sec:appendix:related}).
At a high level, in each design round, these strategies take promising seed antibodies and introduce mutations that are likely to improve properties. Mutations are chosen strategically by incorporating the experimental data collected in previous cycles of \textit{in vitro} validation, often through a trained oracle model.

Most ML strategies for iterative design use only sequence information without structural context, despite such context, such as an experimentally determined antibody-antigen (Ab-Ag) complex, being a valuable source of information.  This is largely because the use of structural context throughout iterative optimization campaigns is challenging: ground-truth structural information is typically only available for the antibody at the starting point of optimization. Indeed, works that do incorporate both sequence and structure \citep{ruffolo2024adapting,luo2022antigen,shanehsazzadeh2023vitro,martinkus2024abdiffuser} only consider single-shot optimization, without incorporating experimental data from multiple rounds of \textit{in vitro} validation into the design process.

In this work, we outline a strategy for embedding sequence-structure generative models into iterative design, actively incorporating laboratory data.
We start by retraining an existing sequence-structure diffusion generative model, DiffAb \citep{luo2022antigen}, which generates antibody complementarity determining regions (CDRs) conditioned on sequence and structure information from an Ab-Ag complex. We propose a pipeline for iterative design such that in a given design round, we use a carefully constructed predicted Ab-Ag complex for the current best molecules as a starting point for generation with this diffusion model. Generated designs can then be ranked with an oracle model trained on experimental data from the optimization campaign, with the best selected for \textit{in vitro} validation. Finally, drawing inspiration from the guided diffusion literature \citep{dhariwal2021diffusion}, we outline a method for more efficient sampling by guiding the diffusion model's denoising process using the oracle model. 

In our experiments, we first demonstrate \textit{in silico} that our method produces distributions of sequences that are enriched for favorable properties, such as strong predicted binding affinity and low predicted polyreactivity. Then, in multiple rounds of \textit{in vitro} validation in a real-world therapeutic optimization campaign, we show that candidates generated by our method both synthesize successfully and are strong binders.

\section{Methods}

\subsection{Background}

\textbf{Notation.} We denote an antibody-antigen (Ab-Ag) complex as \( \mathcal{V} \), where each amino acid in \( \mathcal{V} \) is represented by its residue type \( s_i \in \mathcal{A} = \{\text{A}, \text{C}, \text{D}, \text{E}, \ldots, \text{Y}\} \), $C_{\alpha}$ atom coordinates \( x_i \in \mathbb{R}^3 \), and orientation \mbox{$O_i \in \text{SO}(3)$}. The complex is partitioned into a masked region \( \mathcal{M} \), typically corresponding to the complementarity determining regions (CDRs) to be designed, and an unmasked context \( \mathcal{U} \), such that: $\mathcal{V} = \mathcal{M} \cup \mathcal{U}$. Here, \( \mathcal{M} = \{(s_j, x_j, O_j) \mid j = l+1, \ldots, l+m\} \) represents the masked amino acids, where \( m \) is the number of residues in the CDRs, and \( \mathcal{U} = \{(s_i, x_i, O_i) \mid i \in \{1, \ldots, N\} \setminus \{l+1, \ldots, l+m\}\} \) contains the unmasked context. 
Define $f$ to be an \textit{ oracle} which predicts some property of the antibody from the complex, e.g., its binding affinity to the antigen. 

\textbf{Diffusion Process.} We adopt the formulation from \citet{luo2022antigen}. First, we define a forward diffusion process from $t=0$ to $t=T$. This takes an Ab-Ag complex sampled from the data distribution $p(\mathcal{D})$, denoted $(\mathcal{M}^{(0)}, \mathcal{U})$, and progressively adds noise to the residue type $s$, position $x$, and orientation $O$ of each amino acid in the masked region $\mathcal{M}$ over $T$ steps, resulting in $(\mathcal{M}^{T}, \mathcal{U})$. Further details are in Appendix \ref{sec:appendix:diffusion}. 
Our goal is to learn a reverse diffusion process that takes a noisy input $(\mathcal{M}^{T}, \mathcal{U})$, and iteratively refines it to obtain an estimate $(\widehat{\mathcal{M}^{(0)}}, \mathcal{U})$. As in DiffAb \citep{luo2022antigen}, we achieve this by learning a denoising neural network $G$ that takes a noised complex $(\mathcal{M}^{T}, \mathcal{U})$ and produces three outputs: residue type predictions, denoised coordinates, and denoised orientations. These outputs parameterize the reverse distributions as follows:
\begin{enumerate}[nosep, leftmargin=*]
    \item \textbf{Residue types:}
    \[
    p(s_j^{t-1} \mid \mathcal{M}^t, \mathcal{U}) = \text{Multinomial}\big(G_{\text{type}}(\mathcal{M}^{t}, \mathcal{U})[j]\big),
    \]
    where $G_{\text{type}}(\cdot)$ predicts residue type probabilities for each amino acid $j$ in the masked region $\mathcal{M}$, conditioned on the noisy state $\mathcal{M}^{t}$ and the context $\mathcal{U}$.

    \item \textbf{Positions:}
    \[
    p(x_j^{t-1} \mid \mathcal{M}^{t}, \mathcal{U}) = \mathcal{N}\big(x_j^{t-1}; G_{\text{pos}}(\mathcal{M}^{t}, \mathcal{U})[j], \beta_t I\big),
    \]
    where $G_{\text{pos}}(\cdot)$ predicts the mean of the denoised C$_{\alpha}$ positions, and $\beta_t I$ is the noise variance.

    \item \textbf{Orientations:}
    \[
    p(O_j^{t-1} \mid \mathcal{M}^{t}, \mathcal{U}) = \text{IG}_{\text{SO(3)}}\big(O_j^{t-1}; G_{\text{orient}}(\mathcal{M}^{t}, \mathcal{U})[j], \beta_t\big),
    \]
    where $\text{IG}_{\text{SO(3)}}$ is an isotropic Gaussian distribution over $\text{SO(3)}$, and  $G_{\text{orient}}(\cdot)$ predicts the denoised orientations for each amino acid in $\mathcal{M}$.
\end{enumerate}
The network $G$ is trained to minimize KL divergences between the denoised and true distributions over all timesteps (Appendix \ref{sec:appendix:diffusion}). 
At inference time, the model iteratively denoises a starting complex through repeated applications of refinement and sampling, following the above distributions.

\textbf{Iterative Antibody Design.} In iterative design, the objective is to optimize properties (e.g., binding affinity) of a starting antibody over multiple rounds of optimization. In each round, we do the following: take a set of top-performing antibodies, or seeds, from previous rounds; generate a set of new candidate designs by diversifying the seeds, often using a generative model; select a subset of these candidates to test \textit{in vitro}; screen these in laboratory assays, measuring key properties; and use this data in next round of optimization. Here, we are interested in an iterative design setting where we are provided a starting Ab-Ag complex, and wish to use this to inform our optimization.

\subsection{\textit{DiffAbOpt}: Sequence-Structure Iterative Design}
\label{sec:iterative_design}
\citet{luo2022antigen}, which introduced the DiffAb model, outlined a strategy for single-shot antibody optimization, given access to a trained generative model over Ab-Ag complexes and a starting complex: (1) Apply $t$ steps of noise to locations in the input complex we wish to optimize, following the forward diffusion process;  (2) Denoise the partially noised complex using the trained generative model. By repeating this multiple times, we obtain a set of candidate sequences and structures. 

Here, we outline \textit{DiffAbOpt}: a simple yet effective strategy to extend the above approach to iterative design. In each round of optimization, we apply the single-shot optimization approach to our current set of best performers (rather than the starting antibody) using an estimated Ab-Ag structure for the best performer from a folding model. Concretely, we take each top performer, and then fold its sequence using an antibody folding model \citep[e.g., IgFold][]{ruffolo2023fast} to obtain an estimated structure. We then perform rigid body alignment of the estimated structure to the starter antibody from the original Ab-Ag complex, in order to preserve the original binding pose (we implicitly assume that the binding pose has not substantially changed over the course of optimization).

Once we have these estimated complexes for the best performers, we can then follow the standard approach of applying $t$ steps of noise, and then denoising with our trained generative model, to produce new designs. 
After the top performers are diversified with the above method, we can rank them using a surrogate model that is trained on the property we wish to optimize, using the data acquired so far during optimization. The top ranked variants can then be selected for \textit{in vitro} validation.

\paragraph{Oracle Guidance for Efficient Sampling.}
\label{sec:guidance}
So far, we have incorporated lab data in two ways: selecting top performers, and ranking candidates using a surrogate model trained on the experimental data. 
The denoising sampling process, however, is based purely on the training data distribution $p(\mathcal{D})$. If we wish to optimize a property that has limited representation in $p(\mathcal{D})$, then sampling in this manner may be highly inefficient. Here, inspired by guided sampling in diffusion \citep{dhariwal2021diffusion}, we outline a denoising process that incorporates predictions from the oracle model to generate samples that are biased towards achieving our property optimization goals. 

Our goal is to generate samples from a Product of Experts (PoE) distribution \citep{hinton2002training}, incorporating both the diffusion model and an oracle model $f$, with reweighting factor $\gamma$:
\begin{align*}
p_{\text{PoE}}(\mathcal{M}^{t-1} \mid \mathcal{M}^{t}, \mathcal{U}) \propto  
\ p_{\text{diffusion}}(\mathcal{M}^{t-1} \mid \mathcal{M}^{t}, \mathcal{U})  \exp(f({\mathcal{M}^{t-1}}, \mathcal{U}))^\gamma.
\end{align*}
We make an approximation that the oracle $f$ depends only on the sequence of residue types: $f({\mathcal{M}}^{t}, \mathcal{U}) \approx f(s_{1:N}^t)$. Given that the binding pose is preserved, we find that the oracle model is able to predict properties effectively from this sequence alone.
With this in place, the sampling distributions for orientations and positions remain unchanged, though over multiple timesteps, they are influenced by the guidance through sequence changes. For the sequence of residue types, the sampling is modulated by the oracle guidance term:
$$
p_{\text{PoE}}(\mathbf{s}^{t-1} \mid \mathcal{M}^{t}, \mathcal{U}) \propto p_{\text{diffusion}}(\mathbf{s}^{t-1} \mid \mathcal{M}^{t}, \mathcal{U}) \exp(f(\mathbf{s}^{t-1}))^\gamma,
$$
where $\mathbf{s}^{t}$ is a vector of residue types for the whole sequence.
Sampling from this distribution directly is intractable. 
To address this, and given that the original diffusion model's residue type denoising distribution is already constructed to factorize across sequence indices, we follow \citet{schiff2024simple} and compute the product of experts sampling distribution for each residue location $j$ independently:
\[
p_{\text{PoE}}(s_j^{t-1} = r \mid \mathcal{M}^{t}, \mathcal{U}) \propto p_{\text{diffusion}}(s_j^{t-1}=r \mid \mathcal{M}_t, \mathcal{U}) \exp(f(\tilde{\mathbf{s}}_{j_r}^t))^\gamma,
\]
where $r$ is the amino acid residue type, and $\tilde{\mathbf{s}}_{j_r}^t$ denotes the sequence at timestep $t$ with residue at index $j$ mutated to $r$. This distribution is normalized at each sequence index separately, so is tractable to sample from exactly. 

\begin{figure}[t]
    \centering
    \begin{subfigure}[b]{0.3\textwidth}
        \centering
        \includegraphics[width=\textwidth]{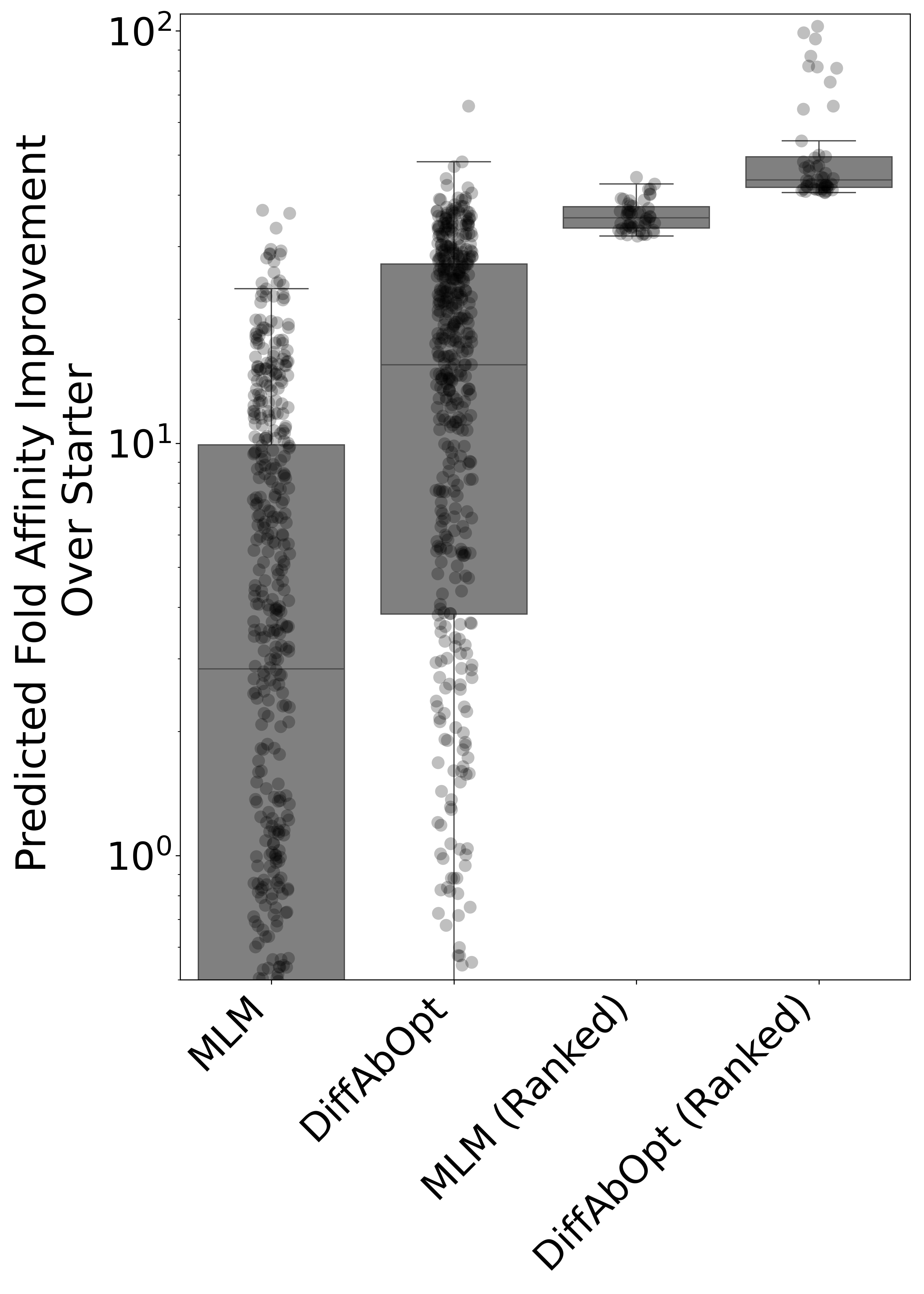}
        \caption{\scriptsize \textit{in silico} evaluation}
        \label{fig:diffabopt-insilico}
    \end{subfigure}
    \begin{subfigure}[b]{0.3\textwidth}
        \centering 
        \includegraphics[width=\textwidth]{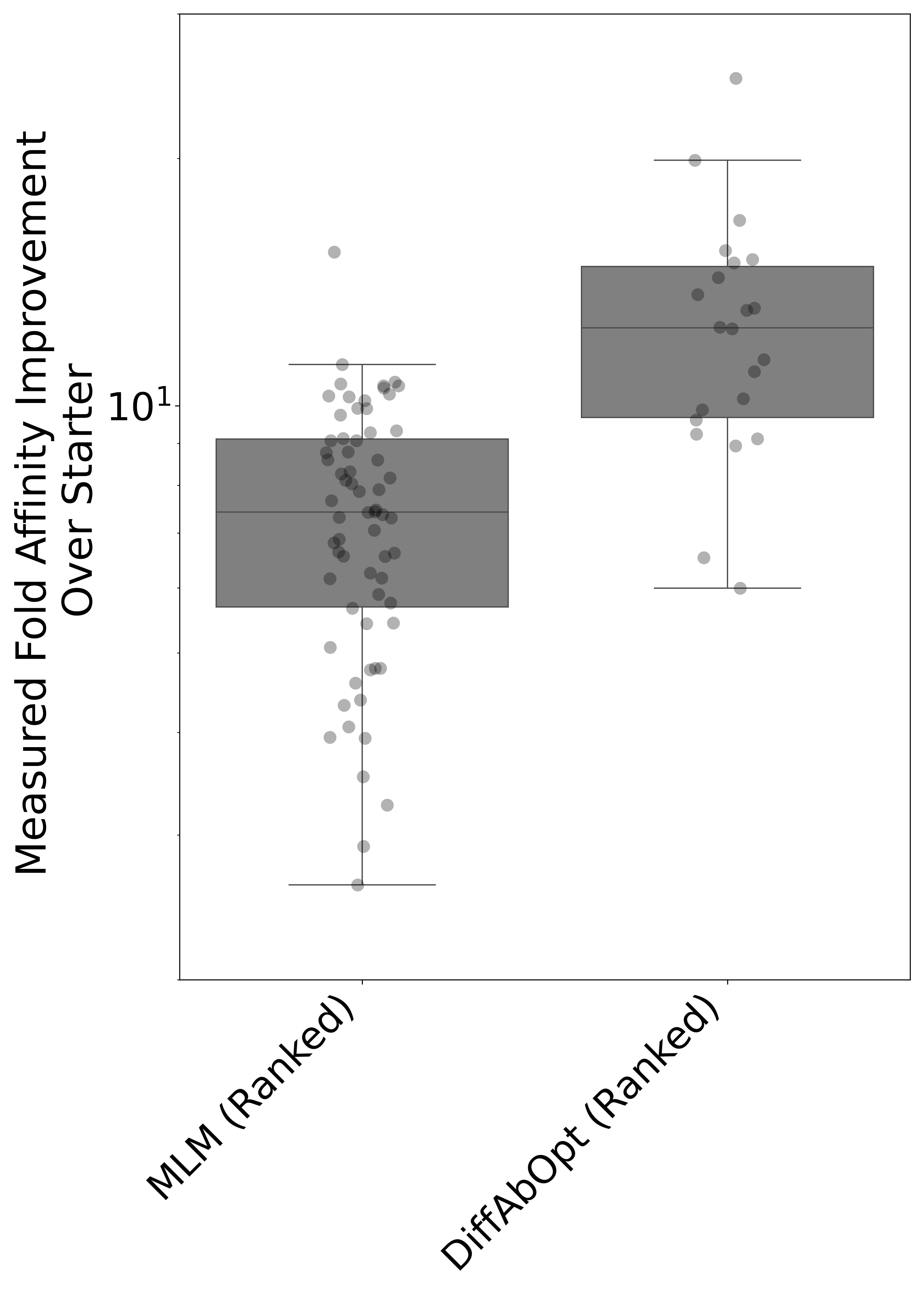}
        \caption{\scriptsize \textit{in vitro}: Mid-stage optimization}
        \label{fig:fig2_invitro_mid}
    \end{subfigure}
    \begin{subfigure}[b]{0.3\textwidth}
        \centering
        \includegraphics[width=\textwidth]{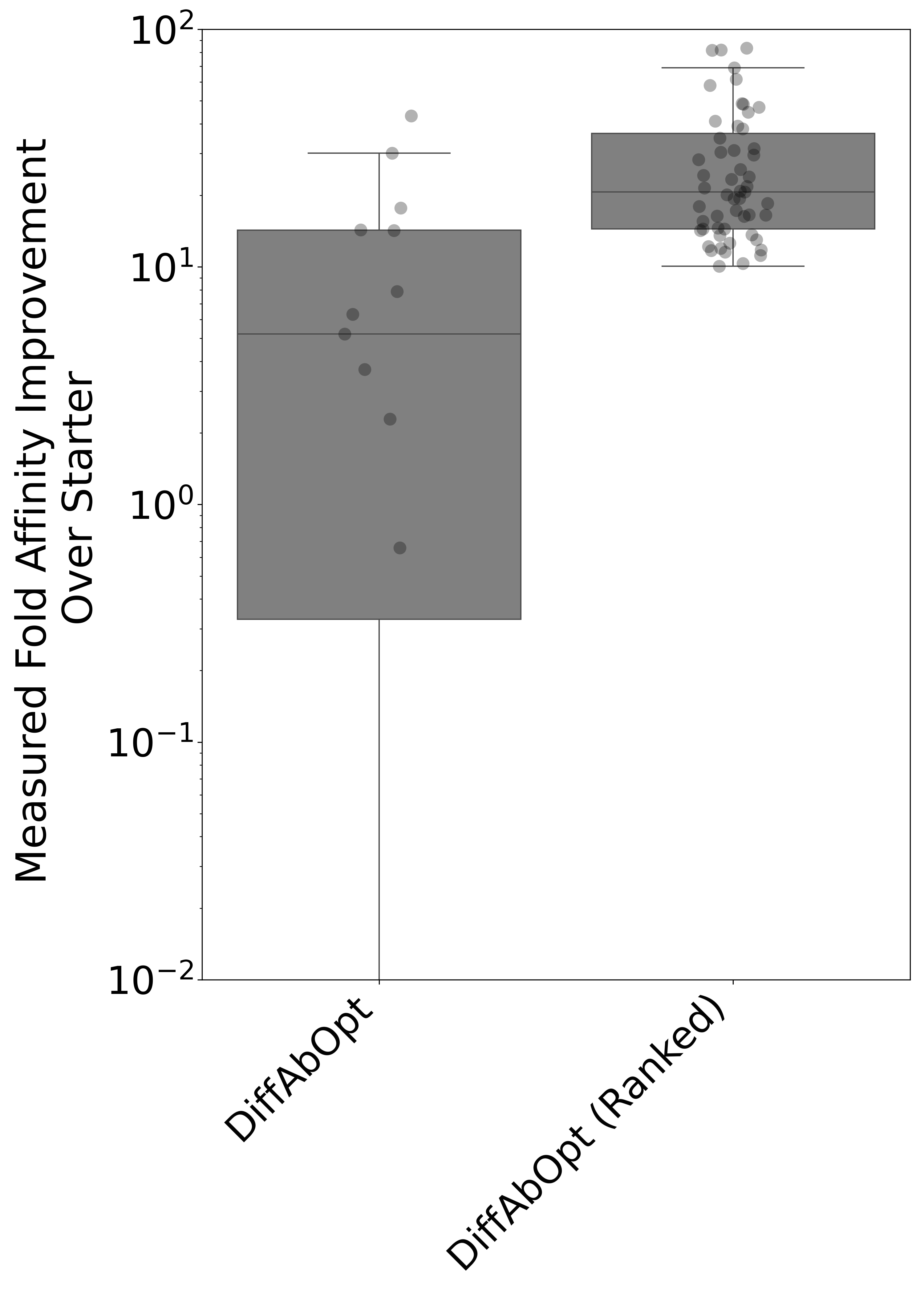}
        \caption{\scriptsize \textit{in vitro}: Late-stage optimization}
        \label{fig:fig2_invitro_late}
    \end{subfigure}
    \caption{\small \textbf{DiffAbOpt is effective both \textit{in silico} and \textit{in vitro}.} In an \textit{in silico} evaluation, DiffAbOpt generates designs that are predicted to have strong affinities, improving over MLM-generated designs. This boost is seen even when we focus on the top-ranked sequences according to an affinity oracle.
    In two \textit{in vitro} evaluations midway through and near the end of the optimization campaign, DiffAbOpt (ranked) outperforms a ranked sequence-only MLM baseline (center) and an unranked DiffAbOpt (right).}
    \label{fig:fig1}
\end{figure}

\section{Experiments and Results}
\subsection{Experimental Setup}
\label{sec:expt_setup}
\textbf{Optimization Campaign.} Our experiments are conducted during a campaign where we were provided a fragment antigen-binding (Fab) antibody with an experimentally determined Ab-Ag crystal structure, and aimed to improve its binding affinity over multiple design rounds.

\textbf{Diffusion model.} We train a sequence-structure diffusion model by adapting the \href{https://github.com/luost26/diffab}{publicly available code} from \citet{luo2022antigen}, making three changes: using the latest SAbDab dataset (with 30\% more data) \citep{dunbar2014sabdab}, training with expanded CDR definitions, and adjusting the residue type denoising posterior. These changes improve the quality of optimized variants (see Appendix \ref{sec:appendix:experiments}).

\textbf{Oracle models.} We train two binding affinity oracles: a CNN ensemble with a ByteNet architecture \citep{yang2022convolutions} for ranking sequences \textit{in silico}; and a lightweight ridge regression model for guided sampling.
We also train a polyreactivity oracle, which is a random forest regressor based on  sequence features from \citet{chen2024human}. Further details are in Appendix \ref{sec:appendix:experiments}. 

\subsection{Evaluating Sequence-Structure Iterative Design}
\label{sec:expts_iterative}
\textbf{\textit{in silico} evaluation.} We select two top-performing antibodies as seeds, which have between 10-15 CDR mutations from the starting antibody. We then generate designs with the following methods, allowing up to 4 CDR edits:
\begin{itemize}[nosep, leftmargin=*]
    \item \textbf{MLM.} Generate 5000 mutants for each seed VH and VL using the Sapiens masked language model (MLM) \citep{prihoda2022biophi} and then combine mutated variable domains. 
    \item \textbf{MLM (Ranked).} Rank the above output with the ByteNet oracle and take the top 50 sequences.
    \item \textbf{DiffAbOpt.} Estimate the seed antibody structure with IgFold \citep{ruffolo2023fast}, and generate 5000 mutants by diversifying each seed through 8 steps of noising/denoising with the diffusion model (without oracle guidance). 
    \item \textbf{DiffAbOpt (Ranked).} Rank the above output with the ByteNet oracle and take the top 50.
\end{itemize}
Figure \ref{fig:diffabopt-insilico} shows the predicted fold affinity improvements for each method.  The DiffAbOpt methods outperform their sequence-only counterparts, with the DiffAbOpt distributions demonstrating enrichment for higher affinity sequences. In Appendix \ref{sec:app:results}, we also provide additional results on the impact of predicted structures on the results.

\textbf{\textit{in vitro} evaluation.} We conduct two \textit{in vitro} evaluations at different stages of the optimization campaign. 
Firstly, we take seeds with 5-10 CDR edits from the starting antibody and design 72 sequences using the MLM (Ranked) method (66 synthesized successfully), and 22 using DiffAbOpt (Ranked) (all 22 synthesized). Figure \ref{fig:fig2_invitro_mid} shows the binding affinities -- we observe that DiffAbOpt (Ranked) outperforms the sequence-only strategy.
Secondly, we take seeds with 15-20 CDR edits and investigate the impact of oracle model ranking: we design 60 sequences using DiffAbOpt (Ranked) (55 synthesized) and 20 using unranked DiffAbOpt (15 synthesized).
Figure \ref{fig:fig2_invitro_late} presents the binding affinities -- we see the clear benefit in ranking designs using the oracle model.

\begin{figure}[t]
    \centering
    \begin{subfigure}[b]{0.3\textwidth}
        \centering
        \includegraphics[width=\textwidth]{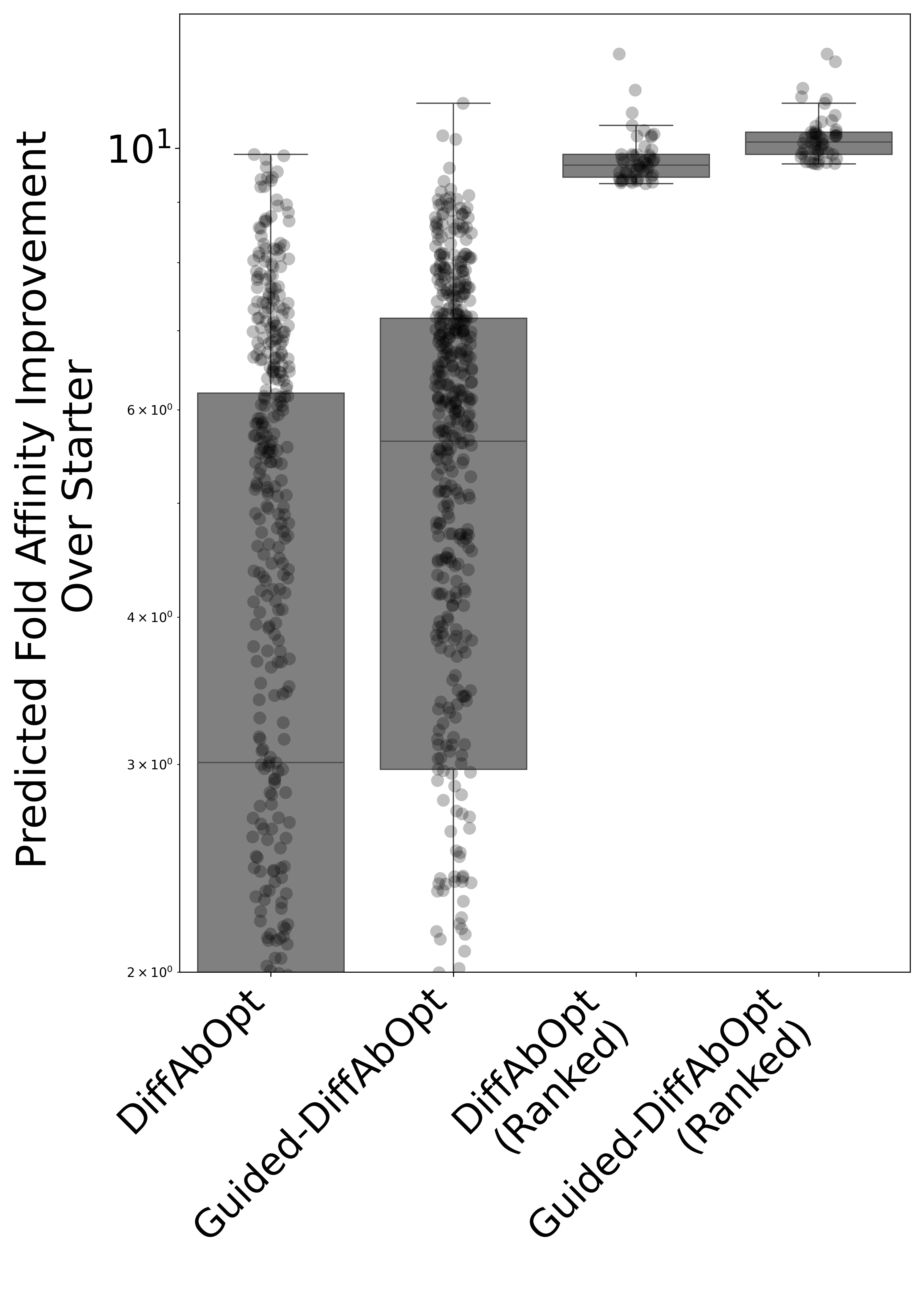}
        \caption{\scriptsize Affinity guidance (\textit{in silico})}
        \label{fig:guidance1-insilico}
    \end{subfigure}
    \begin{subfigure}[b]{0.3\textwidth}
        \centering
        \includegraphics[width=\textwidth]{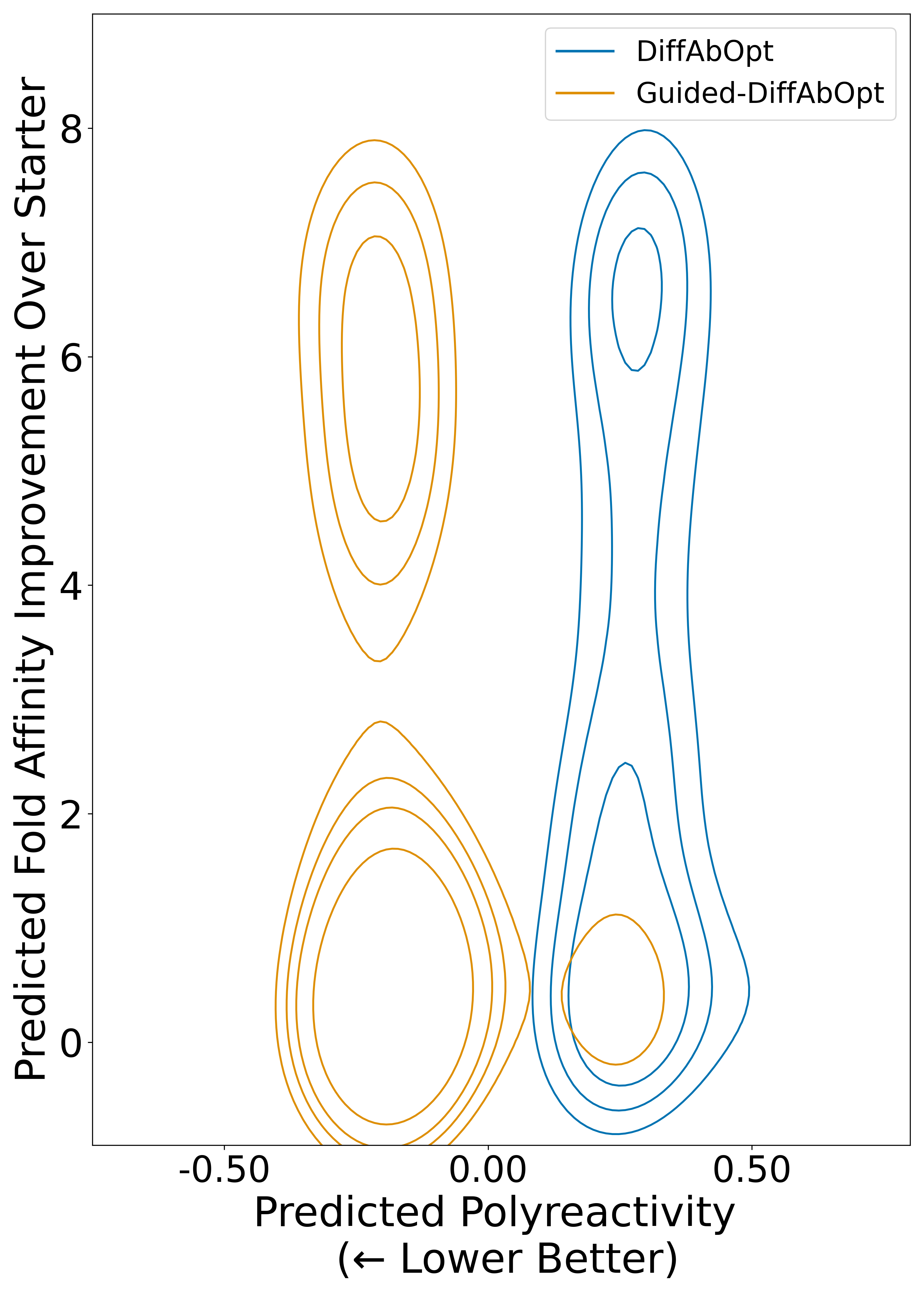}
        \caption{\scriptsize Polyreactivity guidance (\textit{in silico})}
        \label{fig:guidance2-insilico}
    \end{subfigure}
    \begin{subfigure}[b]{0.3\textwidth}
        \centering
        \includegraphics[width=\textwidth]{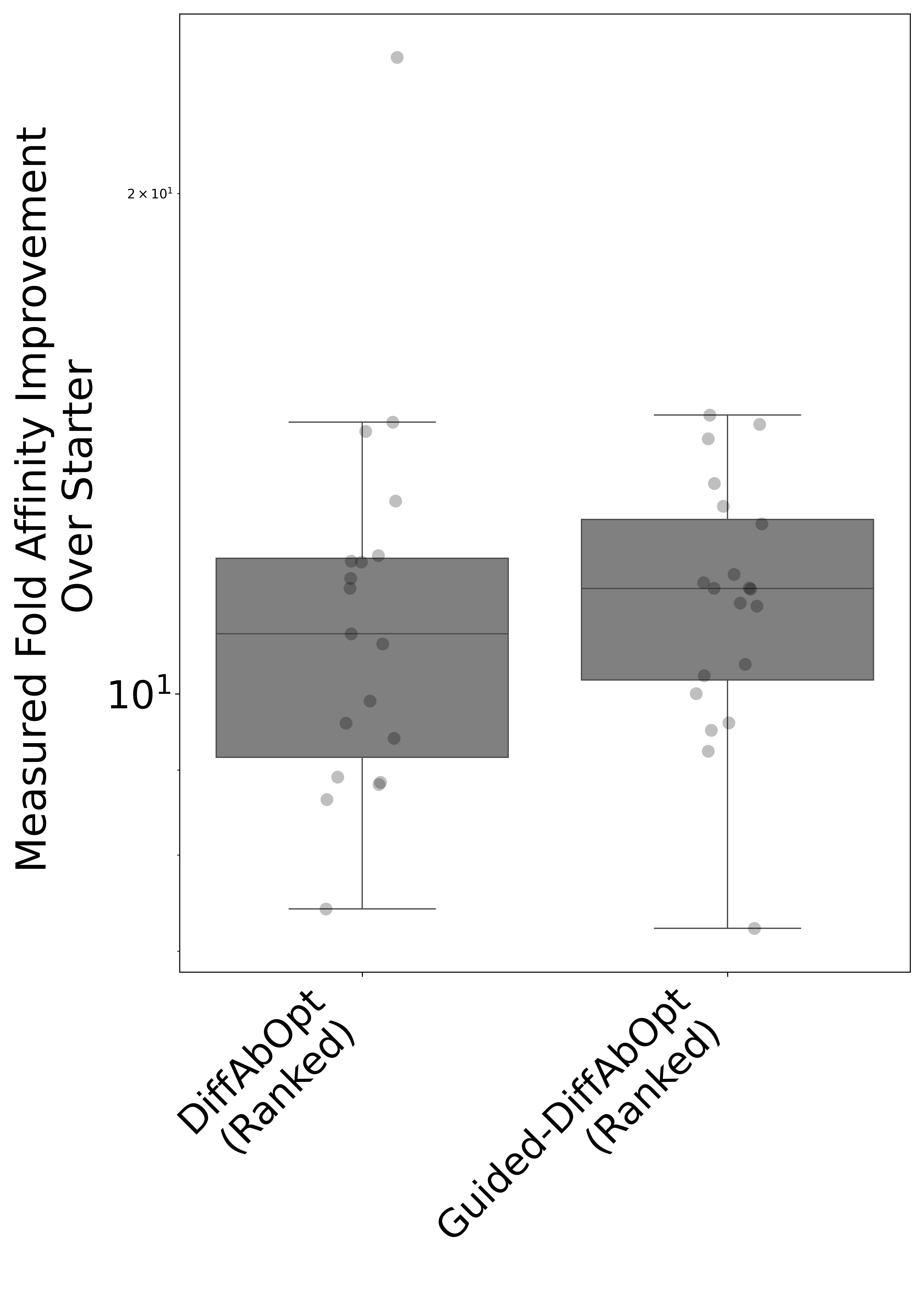}
        \caption{\scriptsize Affinity guidance (\textit{in vitro})}
        \label{fig:guidance1-invitro}
    \end{subfigure}
    \caption{\small \textbf{Guiding DiffAbOpt sampling with oracle models results in designs enriched for desired properties, both \textit{in silico} and \textit{in vitro}}. We train oracle models on data from an optimization campaign and use these models to guide DiffAbOpt sampling. In \textit{in silico} evaluations with two oracles (affinity and polyreactivity, left and center), we find that guidance results in generating more samples with the desired properties. On the right, \textit{in vitro} evaluation of the use of affinity oracle guidance shows a marginal improvement over the unguided approach (though not statistically significant).}
    \label{fig:fig2}
\end{figure}
\subsection{Evaluating Guided Sampling}
\textbf{\textit{in silico} evaluation.} We investigate two oracles: affinity and polyreactivity. We sample 5,000 designs from two seeds (between 10-15 CDR edits from the starting sequence).  Figure \ref{fig:guidance1-insilico} presents results for affinity guidance. In the unranked case, oracle guidance substantially biases the generated sequences towards higher predicted affinities. In the ranked scenario, guidance produces only a small boost in predicted affinity improvements, likely because the base generative model's denoising distribution is already well-enriched for good binders, and the fact that the affinity model used for evaluation is different from the one used at sampling time. Figure \ref{fig:guidance2-insilico} shows the results for polyreactivity guidance. As compared to unguided sampling, using guidance results in a distribution of sequences with lower polyreactivity (preferred) and comparable affinity improvements. This is an example where the samples from the diffusion model are not enriched for sequences with low polyreactivity, and guidance allows more efficient sampling of that region of sequence space. Further results are provided in Appendix \ref{sec:app:results}.

\textbf{\textit{in vitro} validation.} We take two seeds (10-15 CDR edits from the starting sequence) and diversify both using Guided and Unguided DiffAbOpt. We ranked the resulting candidates with the ByteNet oracle and selected the top 20 from each (following developability filtering: see Appendix \ref{sec:appendix:experiments}) for \textit{in vitro} validation. 19 of the unguided samples and 20 of the guided samples synthesized. Figure \ref{fig:guidance1-invitro} shows the affinity results -- we see a marginal boost in the affinity improvements when oracle guidance is incorporated. Increasing the guidance strength during sampling could further improve the affinity gains.

\section{Conclusion}
In this work, we proposed an approach for iterative antibody optimization using sequence-structure diffusion models for antibody-antigen complexes. Our method enables the use of sequence-structure generative models for optimization even when the ground truth structure is available only for the starting molecule. We do this by leveraging \textit{predicted} complex structures of evolving leads as input to the generative model, taking care to preserve the overall binding pose. 
In addition, we describe a strategy to directly incorporate experimental data into the generation process by guiding the diffusion model's sampling distribution with an oracle model trained on data from the optimization campaign. We evaluated our approach \textit{in silico} and in multiple rounds of \textit{in vitro} validation, demonstrating its ability to produce high-affinity binders at different stages of an active antibody campaign.

\subsubsection*{Acknowledgements}
We would like to thank the BigHat Production team, especially Jordan Sullivan, Caleb Ling, Daniel Kim, Eman Khan, Elisabeth Visser, Meg Mora, and Zach Ormsby. We also thank Marcus Rohovie, Emily Delaney, and the rest of the BigHat DS/ML team for productive discussions and insightful suggestions.
\clearpage



\clearpage
\bibliography{iclr2025_conference}
\bibliographystyle{iclr2025_conference}
\clearpage
\appendix
\section{Related Work}
\label{sec:appendix:related}
\paragraph{Sequence-based ML for antibody design.} There has been significant prior work in sequence-based approaches for ML-driven antibody engineering, for instance \citet{stanton2022accelerating, gruver2024protein,amin2024bayesian, maus2022local, zeng2024antibody, freyprotein2024, gordon2024generative}. 
Many of these works take a Bayesian optimization approach to sequence engineering, where an oracle model is fit to experimental data collected during iterative design, and new sequence designs are sampled such that they maximize some acquisition function. This directly captures oracle-predicted performance and predicted uncertainty to trade off exploration and exploitation in a principled fashion. Sequence-based methods naturally allow the incorporation of experimental data from the lab, but are limited in that they are not designed to incorporate structural information, even when available.

\paragraph{Structure-based ML for antibody design.} Prior work in structure-based antibody design includes methods for pure structural design \citep{watson2023novo,krishna2024generalized,bose2024se3}, inverse folding \citep{dauparas2022robust,hoie2023antifold}, and sequence-structure joint design \citep{luo2022antigen,huguet2024sequence,peng2023generative,martinkus2024abdiffuser,shanehsazzadeh2023vitro,campbell2024generative,ruffolo2024adapting,malherbe2024igblend}. Here, we focus on sequence-structure joint design. In terms of antibody optimization, this literature primarily focuses on one-shot structure/sequence modelling, without actively incorporating experimental data from iterative design. This is in part due to the challenge of obtaining high-quality structural information for intermediate lead candidates in an iterative optimization campaign. 

Here, we present a framework for utilizing these structure-based design methods in iterative design, actively incorporating experimental data. This is achieved in three ways: (1) using predicted structures for lead molecules at a given design round as input to generative models, since ground truth experimentally determined structures are likely unavailable for all molecules aside from the starting point of optimization; (2) using oracle models trained on experimental data from the campaign to rank generated designs before selecting a subset for in vitro validation; and (3) directly incorporating oracle models into the generative sampling process to enrich the resulting distribution for designs with favourable properties. 
We choose DiffAb \citep{luo2022antigen} as the sequence/structure generative model in our experiments, since it is publicly available and has been developed in previous work, e.g. \citet{uccar2024benchmarking}.
Compared to other works in this space, we provide data from multiple in vitro validation experiments, demonstrating the promise of our approach.

\paragraph{Guided sampling in diffusion.} Our approach to incorporate oracle models directly into the generative sampling process is inspired by the technique of classifier guidance \citep{dhariwal2021diffusion}.
In this work, we use guided sampling in a discrete diffusion process over sequence-structure data. Several prior works have studied how to formulate guidance in this setting using a range of approaches, including: learning a continuous latent space from a discrete input and applying guidance accordingly \citep{gruver2024protein}; finetuning a base discrete diffusion model to generate samples with high quality, as determined by a guidance model \citep{rector2024steering,wang2024fine}; and finetuning-free approaches by adjusting the sampling distribution used at inference time \citep{nisonoff2024unlocking, schiff2024simple}. In this work, we follow the finetuning-free approach from \citet{schiff2024simple} that assumes a notion of independence among sequence tokens during generation, and incorporate it into a sequence-structure joint diffusion model.

\section{Additional Methods Details}
\subsection{Diffusion Process}
\label{sec:appendix:diffusion}
We follow the formulation of sequence-structure joint diffusion from \citet{luo2022antigen}.
\paragraph{Forward Diffusion.}
The forward diffusion process introduces noise to the data in \( \mathcal{M} \), transitioning from the observed distribution at \( t = 0 \) to a prior distribution at \( t = T \). For each modality, this process is defined as follows:
\begin{enumerate}
    \item \textbf{Residue types:} Residue types \( s_j^t \) are corrupted using a multinomial distribution:
    \[
    q(s_j^t \mid s_j^{t-1}) = \text{Multinomial}\left((1 - \beta_t) \cdot \text{onehot}(s_j^{t-1}) + \beta_t \cdot \frac{1}{20} \cdot \mathbf{1}\right),
    \]
    where \( \beta_t \) is the noise schedule, \( \text{onehot}(s_j^{t-1}) \) converts \( s_j^{t-1} \) to a one-hot encoding, and \( \mathbf{1} \) is a vector of ones. At any timestep \( t \), direct sampling can be written as:
    \[
    q(s_j^t \mid s_j^0) = \text{Multinomial}\left(\bar{\alpha}_t \cdot \text{onehot}(s_j^0) + (1 - \bar{\alpha}_t) \cdot \frac{1}{20} \cdot \mathbf{1}\right),
    \]
    where \( \bar{\alpha}_t = \prod_{\tau=1}^t (1 - \beta_\tau) \).

    \item \textbf{Positions:} The C$_{\alpha}$ coordinates \( x_j^t \) are perturbed using a Gaussian distribution:
    \[
    q(x_j^t \mid x_j^0) = \mathcal{N}\left(x_j^t ; \sqrt{\bar{\alpha}_t} x_j^0, (1 - \bar{\alpha}_t) I\right),
    \]
    where \( \bar{\alpha}_t \) scales the original coordinate, and \( I \) is the identity matrix.

    \item \textbf{Orientations:} Orientations \( O_j^t \) are corrupted using an isotropic Gaussian distribution over \( \text{SO(3)} \):
    \[
    q(O_j^t \mid O_j^0) = \text{IG}_{\text{SO(3)}}\left(O_j^t ; \text{ScaleRot}(\sqrt{\bar{\alpha}_t} O_j^0), (1 - \bar{\alpha}_t)\right),
    \]
    where \( \text{ScaleRot} \) rescales the rotation matrix.
\end{enumerate}

\paragraph{Reverse Diffusion.}
The reverse diffusion process starts from the prior distribution at \( t = T \) and iteratively refines the noisy data in \( \mathcal{M} \) to reconstruct the target distribution, conditioned on the unmasked context \( \mathcal{U} \). We learn a neural network that takes in the masked and unmasked context and produces outputs for the denoised residue types, denoised C$_{\alpha}$ positions, and denoised orientations. For each timestep \( t \), this captures the following posterior distribution:
\begin{enumerate}
    \item \textbf{Residue types:}
    \[
    p(s_j^{t-1} \mid \mathcal{M}^t, \mathcal{U}) = \text{Multinomial}\big(G_{\text{type}}(\mathcal{M}^{t}, \mathcal{U})[j]\big),
    \]
    where \( G_{\text{type}}(\cdot) \) is a neural network that predicts residue type probabilities for amino acid \( j \) based on the noisy state \( \mathcal{M}^t \) and the context \( \mathcal{U} \).

    \item \textbf{Positions:}
    \[
    p(x_j^{t-1} \mid \mathcal{M}^{t}, \mathcal{U}) = \mathcal{N}\big(x_j^{t-1}; G_{\text{pos}}(\mathcal{M}^{t}, \mathcal{U})[j], \beta_t I\big),
    \]
    where $G_{\text{pos}}(\cdot)$ predicts the mean of the denoised C$_{\alpha}$ positions, and $\beta_t I$ is the noise variance.

    \item \textbf{Orientations:}
    \[
    p(O_j^{t-1} \mid \mathcal{M}^{t}, \mathcal{U}) = \text{IG}_{\text{SO(3)}}\big(O_j^{t-1}; G_{\text{orient}}(\mathcal{M}^{t}, \mathcal{U})[j], \beta_t\big),
    \]
    where $\text{IG}_{\text{SO(3)}}$ is an isotropic Gaussian distribution over $\text{SO(3)}$, and  $G_{\text{orient}}(\cdot)$ predicts the denoised orientations for each amino acid in $\mathcal{M}$.
\end{enumerate}

\paragraph{Training Objective.}
The model is trained by minimizing the KL divergence between the forward and reverse processes across all timesteps. The loss over the residue types is:
$$
L_{\text{type}}^t = \mathbb{E}_{\mathcal{M}^t \sim q}\left[\frac{1}{m} \sum_{j=l+1}^{l+m} D_{\text{KL}}\left(q(s_j^{t-1} \mid s_j^t, s_j^0) \parallel \text{Multinomial}\big(G_{\text{type}}(\mathcal{M}^{t}, \mathcal{U})[j]\big)\right)\right],
$$
the loss for the positions is:

$$
\mathcal{L}_{\text{pos}}^t = \mathbb{E}_{\mathcal{M}^t \sim q} \left[\frac{1}{m}\sum_{j \in \mathcal{M}} \left\| x_j^{t-1} - G_{\text{pos}}(\mathcal{M}^{t}, \mathcal{U})[j] \right\|^2 \right],
$$
and the loss for the orientations is:
$$
\mathcal{L}_{\text{orient}}^t = \mathbb{E}_{\mathcal{M}^t \sim q} \left[ \frac{1}{m} \sum_{j \in \mathcal{M}} \left\| (O_j^{0})^\top \hat{O}_j^{t-1} - I \right\|_F^2 \right],
$$
where $\hat{O}_j^{t-1}$ is the output of the orientation predictor $G_{\text{orient}}(\mathcal{M}^{t}, \mathcal{U})$.
The total loss is the weighted sum of the three individual losses, with an expectation taken over all timesteps:
\begin{align}
\mathcal{L} = \mathbb{E}_{t\sim \text{Uniform}(1,T)} \left[\lambda_{\text{type}}\mathcal{L}_{\text{type}}^t + \lambda_{\text{pos}} \mathcal{L}_{\text{pos}}^t + \lambda_{\text{orient}} \mathcal{L}_{\text{orient}}^t \right] \label{eqn:objective}.
\end{align}

\section{Additional Experimental Details}
\label{sec:appendix:experiments}
\subsection{DiffAb Training}
\label{sec:appendix:diffab-training}
As discussed in Section \ref{sec:expt_setup}, we train a sequence-structure generative model by adapting the \href{https://github.com/luost26/diffab}{publicly available code} from \citet{luo2022antigen}. We use the same architecture as \citet{luo2022antigen}, and make three main changes to the training process. 

Firstly, we train on an updated version of the SAbDab dataset (downloaded on 2024/10/17)\citep{dunbar2014sabdab}, with approximately 30\% more data than the dataset used to train the model from \citet{luo2022antigen}. As in the original paper, we remove structures with worse resolution than 4\AA, remove antibodies targeting non-protein antigens, and cluster the antibodies at 50\% sequence identity by HCDR3 sequences to define training/testing sets.

Secondly, we train the model with expanded CDR definitions, rather than the Chothia definition used in the original paper. This expanded definition is based on a union of Chothia, IMGT, and Kabat indices.

Thirdly, we adjust the posterior calculation for the residue type denoising posterior -- we found that a variance term was calculated incorrectly in the original implementation. As a result of this change, we set $\lambda_{\text{type}}=10.0, \lambda_{\text{pos}}=1.0, \lambda_{\text{orient}}=1.0$ in \eqref{eqn:objective}. Without this change, the residue type denoising posterior term in the loss had too small a scale and did not reduce during optimization.

We find that the above changes improve the quality of the sampled variants -- see Appendix \ref{sec:app:results}.

Following the original paper, the model is trained for 200,000 steps with the Adam optimizer \citep{kingma2014adam} and a learning rate of 1e-4. The model is trained to generate the sequence/structure of CDRs given the framework region and antigen.  As with the original paper, we trained the model on patch sizes of 128 residues, based on proximity to anchors adjacent to the CDRs. Further increasing the patch size did not appear to improve generation quality and led to slower training/inference times.

\subsection{Oracle Model Training}
We train oracles on two assays: binding affinity and polyreactivity.

For binding affinity, we train two different oracle models on one-hot sequence encodings:
\begin{itemize}
    \item ByteNet Oracle: This is an ensemble of ten 1D CNNs with a ByteNet/CARP architecture \citep{yang2022convolutions}. We use this model for in silico evaluation, specifically to rank generated sequences and select the top-ranked for promotion to in vitro validation. This model is first pre-trained on in-house Next Generation Sequencing (NGS) datasets with around 75k sequences, and then fine-tuned on binding affinity datasets using all available data from the optimization campaign at the specific design round (ranging from 4k to 8k sequences). 
    \item Ridge Regression Oracle: this is a lightweight affinity predictive model that is used in guided sampling during denoising. It is trained directly on all the available binding affinity data from the optimization campaign at the specific design round. We opt to use this during guided sampling instead of the ByteNet oracle since it is a smaller model and faster to run inference on.
\end{itemize}

For polyreactivity, we train a random forest regressor based on handcrafted sequence features from \citet{chen2024human}. This is trained on a dataset of ~250 sequences, all of which have flow cytometry polyreactivity measurements.

\subsection{Sampling Hyperparameters and Implementation}
\label{sec:appendix:sampling_hyper}
\paragraph{Edit distance.} For all in silico and in vitro evaluations, we allow up to 4 CDR edits (across all VH and VL CDRs). This is to ensure a reasonable rate of synthesis and binding in the generated dataset.

\paragraph{DiffAbOpt.} We experimented with various sampling configurations in terms of (a) what CDRs have noise applied to them, and (b) how many steps of noise to apply. We experimented with noising all subsets of CDRs for 2, 4, 8 and 16 steps and denoising for the same number of steps as was used during noising. 
We found that for a given sampling budget (total number of generated samples, prior to edit distance filtering), we obtained the best results by applying 8 steps of noise to all six CDRs simultaneously and applying denoising for 8 steps. With more steps of noising, the model generates very few sequences within the required maximal edit distance. Denoising different subsets of the CDRs did not appear to result in much change over noising/denoising them all together.

\paragraph{Improving the Efficiency of Guided Sampling.} In order to improve the efficiency of guided sampling, we found it valuable to cache oracle predictions for a given sequence, and retrieve from the cache instead of predicting. 

\paragraph{Formulating Guidance.} For the affinity oracle, we set the guidance reweighting factor $\gamma = 2$. For the polyreactivity oracle, given that lower polyreactivity is better and due to the natural scale of the model's outputs, we set $f_{\text{guidance}}(\mathbf{s})= -f_{\text{polyreactivity}}(\mathbf{s})$ and set the reweighting factor $\gamma = 10$.
\begin{figure}[t]
    \centering
        \includegraphics[width=0.6\textwidth]{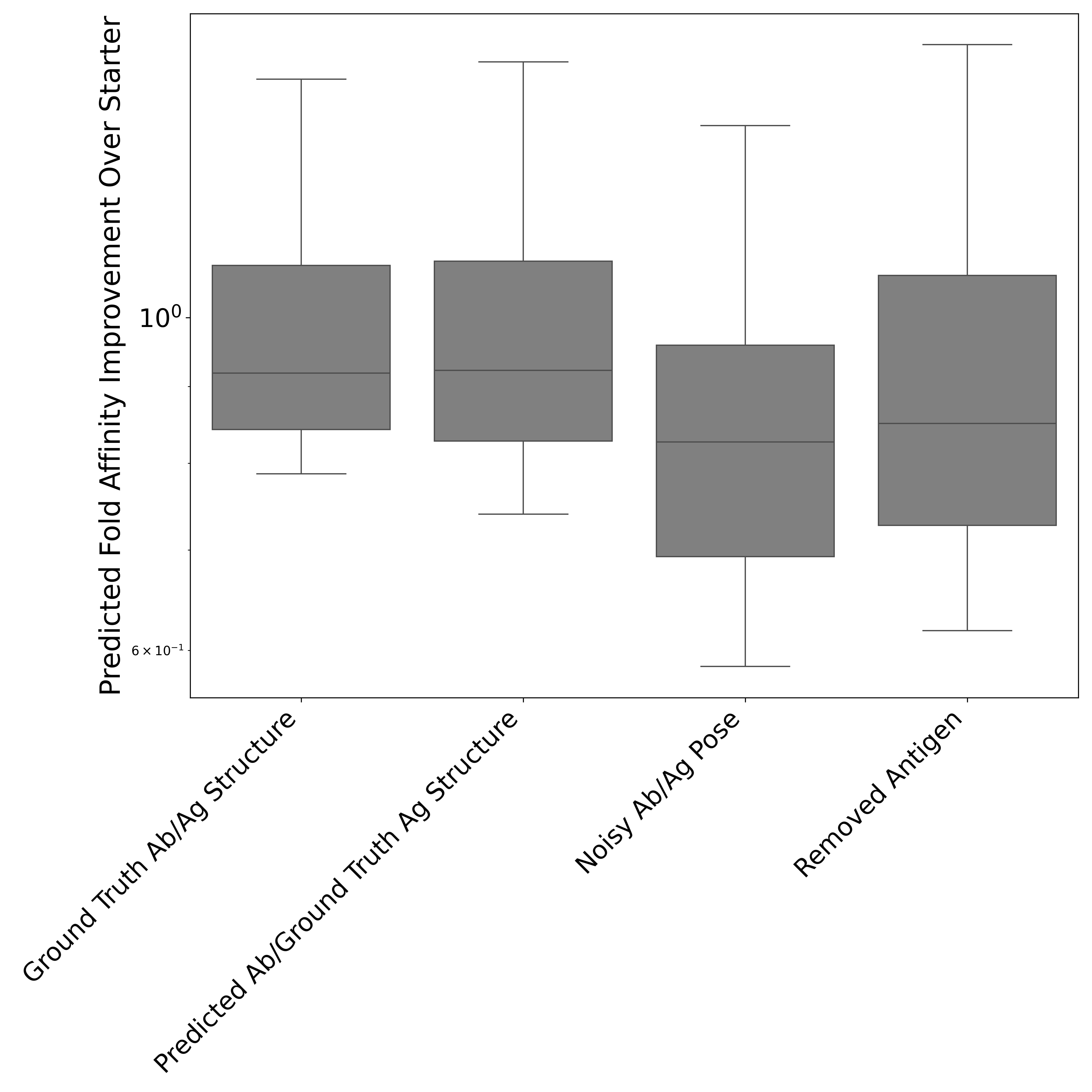}
        \caption{\small \textbf{Investigating the effect of structural noise in the distribution of top-ranked generated sequences.} We compare how the distribution of the top ranked 500 designs vary when we add different kinds of noise to the structure taken in as input by the diffusion model. As a seed, we use the antibody at the starting point of the optimization campaign, for which we have an experimentally determined structure. The leftmost box uses the ground-truth experimentally determined structure of the Ab/Ag complex. The second box uses a predicted structure for the Ab, which shows increased variance relative to using the ground truth structure, but a similar median performance. Next, we apply noise to the structure by applying a rigid body rotation to the antibody (third box), finding that the resulting sequences are predicted to perform substantially worse. Finally, we try removing the antigen altogether (rightmost box), which is better than a noisy pose, but worse than using the ground truth structure or the predicted Ab structure.}
        \label{fig:diffabopt-structure-ablation}
\end{figure}

\subsection{Additional Results}
\label{sec:app:results}
\paragraph{Impact of predicted structures.} Here, we explore how different kinds of structural noise impact the distribution of predicted affinities of the sequences generated by the diffusion model. For this experiment, we use the antibody at the starting point of the optimization campaign as the seed, since it has a ground truth empirically determined Ab/Ag crystal structure. Figure \ref{fig:diffabopt-structure-ablation} presents the results, showing the distribution of predicted fold affinity improvements of the top 500 ranked designs. We observe that using the ground truth structure and predicted Ab structure (left two boxes) perform similarly, with the ground truth structure showing less variance in predicted performance. Adding noise to the Ab/Ag pose (via rigid body rotation of the antibody) and removing the antigen altogether (right two bars) both worsen performance. This suggests that minor changes, such as using a predicted antibody structure but preserving the same pose, are well-tolerated in terms of the best-quality designs that the model can produce.

\paragraph{Impact of training changes: extra data, CDR definitions, and posterior.}
\begin{figure}[t]
    \centering
        \includegraphics[width=0.6\textwidth]{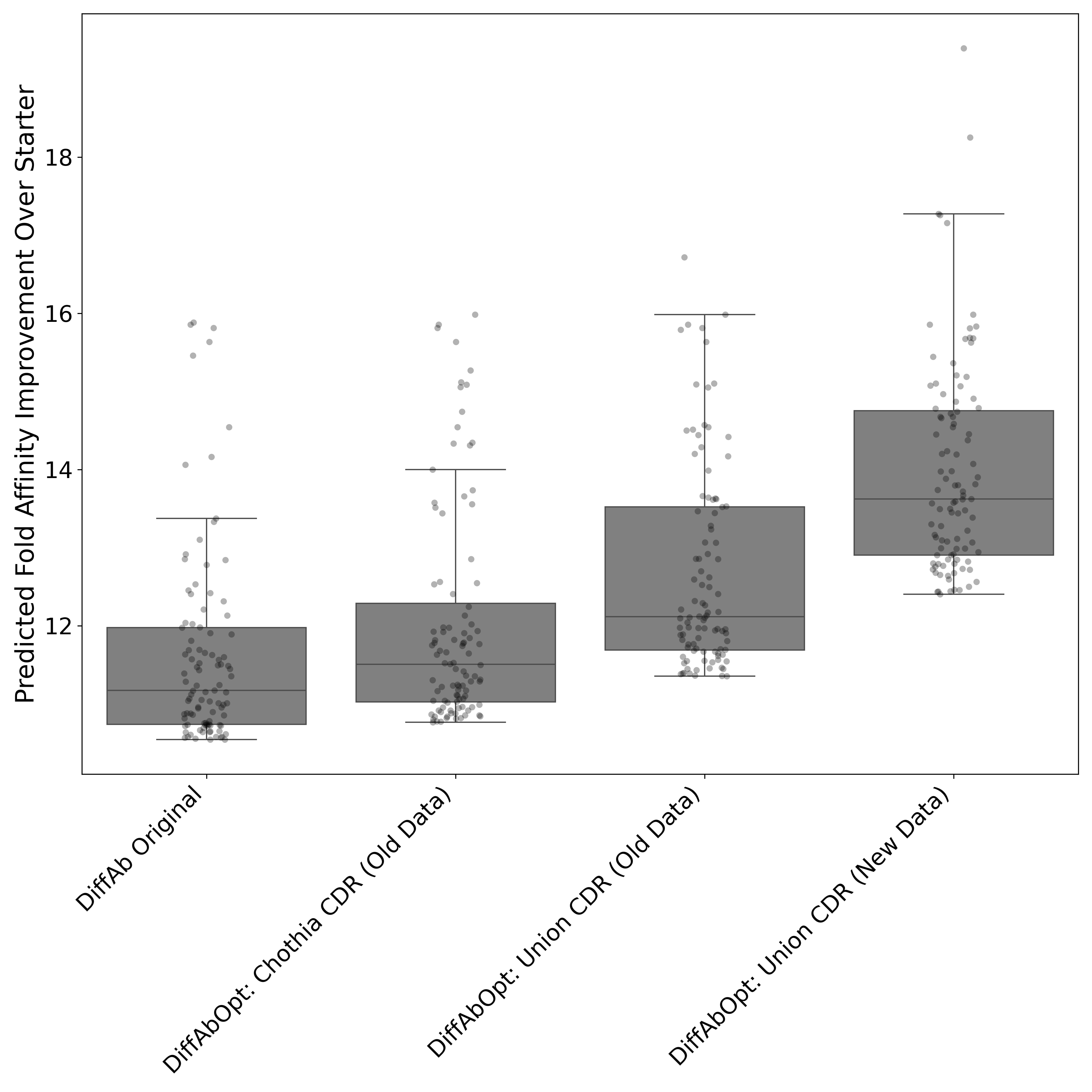}
        \caption{\small \textbf{Expanded CDR definitions and training on larger datasets boosts predicted performance of generated variants.} We compare the distribution of predicted affinity improvements from top-ranked sequences generated from different DiffAb models. From left to right, these are: the original model from \citet{luo2022antigen}, a model trained with an adjusted sequence posterior,  a model trained with expanded CDR definitions and the adjusted posterior, and finally a model trained on a more recent dataset version with adjusted CDR definitions and the adjusted posterior. We observe boosts in the quality of generated sequences when we combine these changes.}
        \label{fig:diffabopt-datacdr-comparison}
\end{figure}
Figure \ref{fig:diffabopt-datacdr-comparison} shows the predicted fold affinity improvements for different DiffAb models, focusing on the top 100 ranked variants for each. (Note that ``Ranked'' has been excluded in axis labels to reduce clutter). For all of these, we followed the sampling hyperparameters discussed in Section \ref{sec:appendix:sampling_hyper}.

The leftmost box has sequences from the DiffAb model provided with the published paper by \citet{luo2022antigen}. Immediately to the right, we have generations after fixing the posterior calculation, which has a marginal improvement. Next, we incorporate the expanded CDR definitions into training and observe a clear boost in performance. Finally, the rightmost box has the posterior fix, expanded CDR definitions, and training on a larger dataset, again showing a boost in performance. This benefit of more data was also seen in \citet{uccar2024benchmarking}, suggesting that expanding the data set further could improve the quality of the generation even more.

\begin{figure}[t]
    \centering
    \begin{subfigure}[b]{0.4\textwidth}
        \centering
        \includegraphics[width=\textwidth]{figures/polyrx_guidance_comparison_dr48_seeds_only_unranked.png}
        \caption{\scriptsize Polyreactivity guidance without affinity ranking}
        \label{fig:app:guidance-unranked}
    \end{subfigure}
    \begin{subfigure}[b]{0.4\textwidth}
        \centering
        \includegraphics[width=\textwidth]{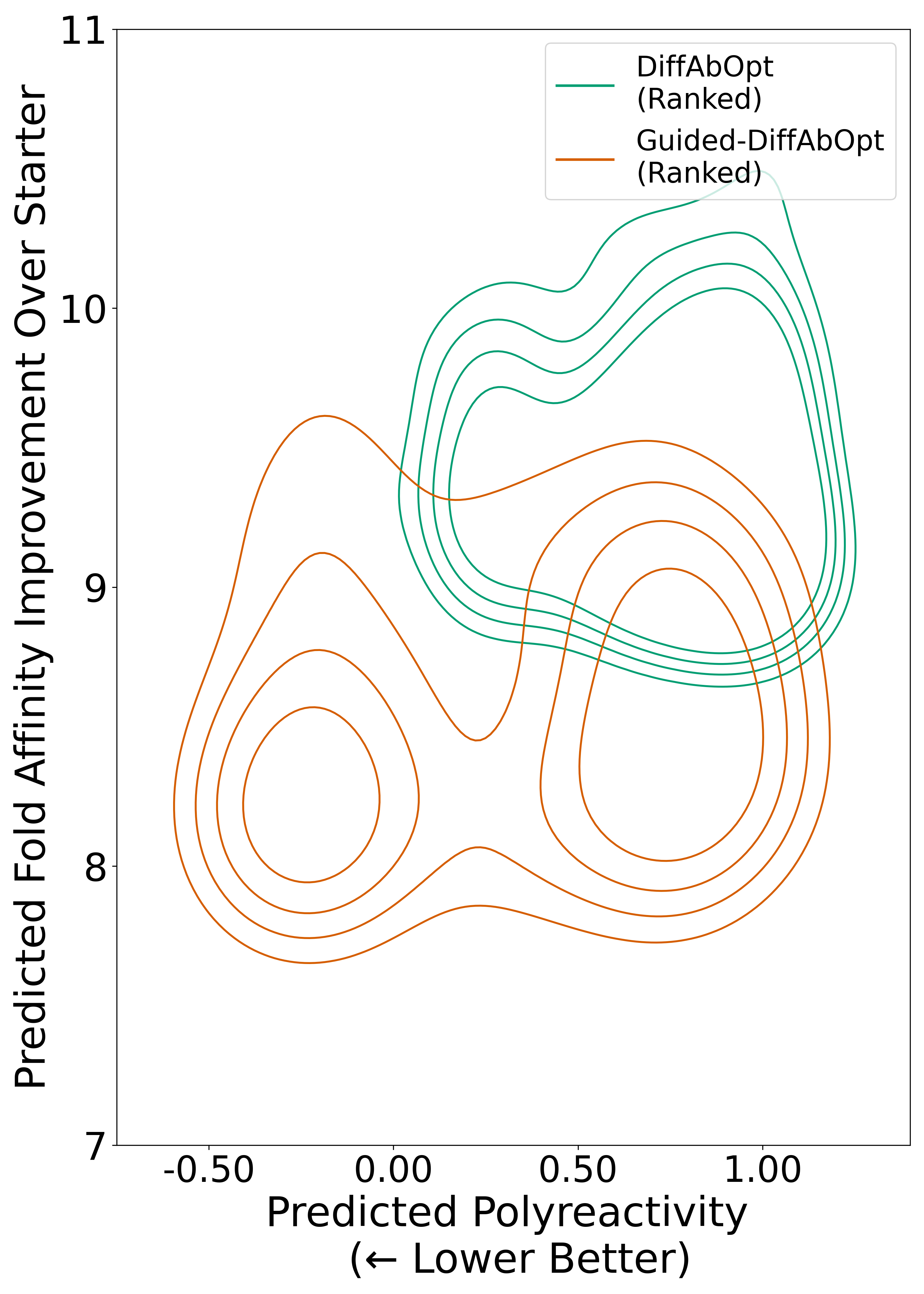}
        \caption{\scriptsize Polyreactivity guidance with affinity ranking}
        \label{fig:app:guidance-ranked}
    \end{subfigure}
    \caption{\small \textbf{Studying the impact of polyreactivity guidance with and without affinity model ranking}. We investigate the impact of polyreactivity guidance without (left) and with (right) affinity oracle ranking post generation, showing the distribution of the top 100 ranked sequences. We see that among this top ranked set, we are able to generate sequences with lower polyreactivity while also maintaining comparable boosts in affinity.}
    \label{fig:app:guidance}
\end{figure}

\paragraph{Further results on polyreactivity guidance.} In the main paper, we showed that polyreactivity guidance resulted in generated sequences that had lower polyreactivity (desirable) and comparable predicted affinity improvements. Here, we investigate taking generated sequences, with and without polyreactivity guidance, and applying post-hoc ranking with the ByteNet affinity oracle. 
Due to the differing distribution shapes between unranked and ranked sequences, visualizing this on one KDE plot is challenging, so Figure \ref{fig:app:guidance} compares results with/without ranking on two separate plots. As with the unranked evaluation, we see that even after ranking with the oracle model (taking the top 100 ranked designs), we see comparable predicted affinity improvements to unguided sampling, while still observing clear reductions in the resulting predicted polyreactivity. These two objectives can directly be at odds \citep{chen2024human}, so being able to trade them off in generation is useful in practice.

\subsection{in vitro validation details}
After generation and ranking, sequences were filtered based on key developability criteria (e.g., isoelectric point, excluding known sequence liabilities) before being selected for wet lab screening. Sequences were synthesized in Cell-Free Protein Synthesis (CFPS) as scFvs and binding affinities of those that synthesized in adequate quantities and purity were measured via BLI (Octet).

\end{document}